\documentclass[sigconf]{acmart}
\copyrightyear{2022} 
\acmYear{2022} 
\setcopyright{acmcopyright}\acmConference[WSDM '22]{Proceedings of the Fifteenth ACM International Conference on Web Search and Data Mining}{February 21--25, 2022}{Tempe, AZ, USA}
\acmBooktitle{Proceedings of the Fifteenth ACM International Conference on Web Search and Data Mining (WSDM '22), February 21--25, 2022, Tempe, AZ, USA}
\acmPrice{15.00}
\acmDOI{10.1145/3488560.3498458}
\acmISBN{978-1-4503-9132-0/22/02}

\usepackage[misc]{ifsym}
\usepackage{amsthm}
\usepackage{graphics}
\usepackage{subfigure}
\usepackage{graphicx}
\usepackage{balance}
\usepackage{amsmath,amsfonts}
\usepackage[T1]{fontenc}

\theoremstyle{definition}
\newtheorem{definition}{Definition}
\newcommand{\ours}{Triangle Graph Interest Network}
\newcommand{{\short}}{TGIN}

\usepackage{color}

\begin{document}
\fancyhead{}
\title{Triangle Graph Interest Network for Click-through \\ Rate Prediction}


\author{Wensen Jiang\textsuperscript{1}*,
Yizhu Jiao\textsuperscript{2}*, 
Qingqin	Wang\textsuperscript{2}, 
Chuanming Liang\textsuperscript{1},
Lijie Guo\textsuperscript{1},
Yao	Zhang\textsuperscript{2}, \\
Zhijun Sun\textsuperscript{1},
Yun Xiong\textsuperscript{2,3}\textsuperscript{(\Letter)}, 
Yangyong Zhu\textsuperscript{2,3}}
\affiliation{
\institution{
{\textsuperscript{1}Alibaba Group, China} \\
{\textsuperscript{2}Shanghai Key Laboratory of Data Science, School of Computer Science, Fudan University, China} \\
{\textsuperscript{3}Shanghai Institute for Advanced Communication and Data Science, Fudan University, China} \\}
{\textsuperscript{1}wensen.jws@alibaba-inc.com, \textsuperscript{2}\{yzjiao18, qqwang18, yunx\}@fudan.edu.cn } \\
\thanks{* denotes equal contributions. \\ \textsuperscript{\Letter} denotes corresponding author. }
\country{}
}

% \author{Anonymous authors}
% \affiliation{%
%   %\institution{Paper under double-blind review}
% }

\newcommand{\commentzy}[1]{{\color{red}[ZY: {#1}]}}
% \newcommand{\commentzy}[1]{}

\renewcommand{\shortauthors}{Jiang and Jiao, et al.}

\begin{abstract}
    Click-through rate prediction is a critical task in online advertising. 
    Currently, many existing methods attempt to extract user potential interests from historical click behavior sequences. 
    However, it is difficult to handle sparse user behaviors or broaden interest exploration. 
    Recently, some researchers incorporate the item-item co-occurrence graph as an auxiliary. 
    Due to the elusiveness of user interests, those works still fail to determine the real motivation of user click behaviors. 
    Besides, those works are more biased towards popular or similar commodities. 
    They lack an effective mechanism to break the diversity restrictions. 
    
    In this paper, we point out two special properties of triangles in the item-item graphs for recommendation systems: Intra-triangle homophily and Inter-triangle heterophiy. 
    Based on this, we propose a novel and effective framework named {\ours} ({\short}).
    For each clicked item in user behavior sequences, we introduce the triangles in its neighborhood of the item-item graphs as a supplement.  
    {\short} regards these triangles as the basic units of user interests, which provide the clues to capture the real motivation for a user clicking an item. 
    We characterize every click behavior by aggregating the information of several interest units to alleviate the elusive motivation problem. 
    The attention mechanism determines users' preference for different interest units. 
    By selecting diverse and relative triangles, {\short} brings in novel and serendipitous items to expand exploration opportunities of user interests. 
    %As the result, our proposed method can break diversity restrictions. 
    Then, we aggregate the multi-level interests of historical behavior sequences to improve CTR prediction. 
    Extensive experiments on both of public and industrial datasets clearly verify the effectiveness of our framework.

\end{abstract}



\maketitle


\section{Introduction}

%$$
%\operatorname{Salience}(e)=\left(1+\log (freq(e))^{2}\right) \log \left(\frac{N_{-} bs}{bsf(e)}\right), 
%$$
%where freq $(e)$ is the frequency of event $e, N_{-}bs$ is the number of background documents,  and $\operatorname{bs} f(w)$ is the background document frequency of event $e$.

Click-Through Rate (CTR) prediction is an essential task for recommendation systems in industrial applications, such as online advertising and sponsored search, etc \cite{covington2016deep}. 
Currently, existing works \cite{song2016multi, covington2016deep, yu2016dynamic, zhou2018deep, zhou2019deep, zhou2018atrank, feng2019deep, li2019graph, xiao2020deep} extract user interests just from historical click behavior sequences to improve the performance of this task. 
However, it is difficult to handle sparser user behaviors or jump out of their specific behaviors for possible interest exploration \cite{li2019graph}.

Recently, a few graph embedding methods are suggested to alleviate these problems \cite{wang2018billion, li2019graph, feng2020mtbrn}. 
They introduce item-item co-occurrence graphs as an auxiliary to reveal the hidden patterns of user historical behaviors.
However, the existing works emphasize discovering user interest from historical clicked item sequences, which causes the problem of \textbf{elusive motivation}.
In general, user interests are implicit and elusive while each item has multiple attributes. 
It is difficult to determine the real motivation for a user clicking an item. 
For example, users may click on a skirt because of its color, material, or style. 
Yet, related works lack an effective mechanism to capture implicit user interests.
In addition, graph embedding based methods attempt to aggregate neighborhoods in the item-item graph to enrich item representations and broaden the exploration. 
Nevertheless, they still suffer from the problem of \textbf{diversity restriction}. 
For a specific item, most of its neighbors in the item-item graph are similar (such as the case shown in Figure 1). 
Although some works adopt the weight-based strategy to sample the neighbors with higher weights, it is biased towards the popular commodities. 
Hence, it is a challenge to make diverse recommendations beyond user existing interests.
 
%Explore with more interest. 
To alleviate these problems, we introduce the triangles in the item co-occurrence graph as the basic units of user interests. 
Triangle is an important structure in complex graph analysis in different fields, such as social networks \cite{watts1998collective, newman2002random, leskovec2010predicting} and protein networks \cite{leskovec2010predicting}. 
Many existing works in graph theory have mentioned the properties of triangles \cite{wasserman1994social, wimmer2010beyond}. 
For instance, the homophily of triangles indicates the nodes in a triangle are often relatively similar \cite{kolountzakis2012efficient}.
Statistics and analysis reveal two special properties of triangles in item co-occurrence graphs for recommendation systems.
1) \textbf{Intra-triangle homophily}. The three items in a triangle usually share some common attributes. 
It can reflect user's real motivations to click these items.
Therefore, triangles can be regarded as the most basic user interest unit. 
Compared with modeling single items, modeling triangles is more helpful to capture users' elusive and implicit interests. 
Moreover, item-item co-occurrence graphs in real-world scenarios are usually large-scale and inevitably noisy. 
By modeling triangles, we can filter out isolated items and reduce noises to improve the quality of user interest extraction.
2) \textbf{Inter-triangle heterophily}. The shared attributes of different triangles are distinct. 
They can reflect different aspects of user interests.
A variety of triangles can introduce novel and diverse commodities to users. 
Consequently, users can obtain more exploration opportunities and expanding their interests. 
For example, we can recommend to users not only clothes of the similar material or color, but also other products of the same store or brand.

 \begin{figure}[!tbp]
     \centering
     \includegraphics[width=0.8\linewidth]{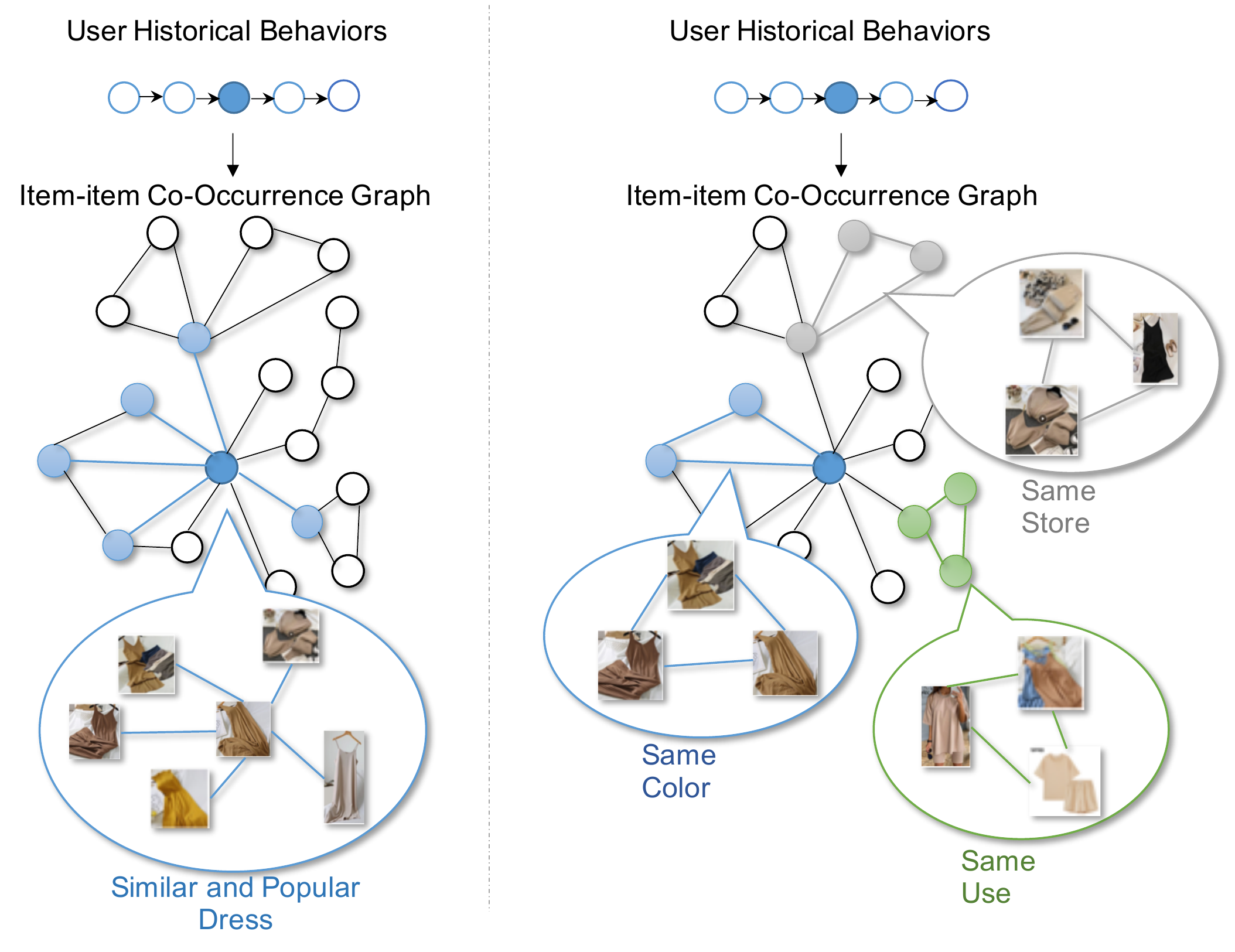}
     \vspace{-10pt}
     \caption{Case Study. Existing work is biased towards the similar and popular items in the left figure. Our method introduces different triangles as diverse interest units, such as skirts of the same color, store and use in the right figure.}
     \label{fig:task}
     \vspace{-10pt}
 \end{figure}

Triangle structures widely exist in graph structures.
To better capture and broaden user interests, the essential concern is how to choose the triangles for the recommendation system.
The following three principles need to be considered.
1) Relevance. To make better use of the user's historical click behavior, the sampled triangle structure should have a strong correlation with the item being clicked. 
2) Diversity. To further expand user interests, it is required that the sampled triangles are as different and diverse as possible. 
3) Efficiency. The CTR prediction task has a high requirement for timeliness. Thus, we need to consider efficient extraction and selection strategies. In addition, limiting the number of triangles makes it more feasible.

In this paper, by integrating above main ideas together, we propose a novel, effective and scalable CTR prediction model, {\ours} ({\short}) \footnote{Source code: \href{https://github.com/alibaba/tgin}{https://github.com/alibaba/tgin}}. 
Considering the homophily of triangles in the item-item co-occurrence graph, our method takes advantage of triangle structures as the basic units of user interests to model complex user behaviors and alleviates the elusive motivation problem. 
Instead of relying on single clicked items, our method can reduce the noise caused by complicated user behaviors, and capture implicit user interests effectively. 
We design an efficient triangle extraction and selection mechanism. 
In the item-item graph, a set of relevant triangles is extracted from the neighborhood of the clicked items, which provides clues to potential user interests. 
We adopt the selection mechanism to choose a few triangles with diversity to bring in serendipitous items for users and expand exploration opportunities. 
We represent every user click behavior by aggregating the information of several interest units. 
The attention mechanism determines the user's preference for different interest units. 
As the result, our proposed method can break diversity restrictions. 
Based on this, we model the historical behavior sequences including user clicks at different times to learn the user's profile, which can better capture the trend of user interests over time.

The main contributions of this paper are summarized as follows.
\begin{itemize} 
\item We propose a novel and effective model, Triangle Graph Interest Network, which introduces triangle structures in item-item co-occurrence graphs as the basic unit of user interests. 
Compared with modeling single clicked items, our method can capture the implicit user interests effectively.
\item We design an efficient triangle extraction and selection mechanism to produce relevant and diverse interest units. 
It can provide exploration opportunities to discover serendipitous items. 
Besides, the attention mechanism can determine users' preferences and broaden their interests.
\item In the experiments, our method significantly outperforms the state-of-the-art baselines on both public and industrial datasets. 
\end{itemize} 

%The organization of the remaining parts of this paper is as follows. Section 2 introduces some related work. Section 3 gives a detailed description of our model. Section 4 presents our experimental results and analyses on both public and industrial datasets. 

\section{Preliminary}
In recommender systems, there are a series of historical clicks between users and items typically. Let $\mathcal{U}$ denote a set of users and $\mathcal{I}$ denote a set of items. We denote user clicks as $\mathcal{R}=\{(u, i, \mathcal{B}_{u}, y) \mid u \in \mathcal{U}, i \in \mathcal{I}\}$. Here, $\mathcal{B}_u \subset \mathcal{I}$ represents historical behaviors ( i.e., item list) for user $u$. 
$y \in\{0,1\}$ is the implicit feedback when recommending item $i$ to user $u$, where $y=1$ when user $u$ clicking item $i$ is observed, and $y=$ 0 otherwise.

\begin{definition} (Item-Item Co-Occurrence Graph): In this paper, the item-item co-occurrence graph is defined as an undirected graph $\mathcal{G}=\{\mathcal{V}, \mathcal{E}\}$, where $\mathcal{V}$ is the node set and $\mathcal{E}$ is the edge set. Each node in $v \in \mathcal{V}$ denotes an item. For user $u \in \mathcal{U}$, if two items $v_0$ and $v_1$ co-occur in his historical behavior sequence $\mathcal{B}_{u}$, then the triplet $(v_0, v_1, w) \in \mathcal{E}$. $w$ denotes the co-occurrence times. 
\end{definition}

In practice, to construct an item-item co-occurrence graph, we set a sliding window and slide it on behavior sequences of all users. 
Each pair of items within the window are connected by an undirected edge.
The weight of each edge is assigned as the total number of occurrences of the two connected items in all users' behaviors. 

\begin{definition} (Triangle): In an item-item co-occurrence graph $\mathcal{G}=\{\mathcal{V}, \mathcal{E}\}$, for node $v \in \mathcal{V}$, we denote the set of its K-order (K>0) neighbor nodes as $\mathcal{V}_v$.
For three nodes $v_0, v_1,$ and $v_2 \in \mathcal{V}_v$, we define the triplet $tr = (v_0, v_1, v_2)$ as a triangle only if $(v_0, v_1), (v_0, v_2),$ and $(v_1, v_2) \in \mathcal{E}$. 
\end{definition}

The number of triangles can increase exponentially with the number of nodes. Thus, to effectively incorporate triangle structure information, we set $K$ as a relative small number. 
Moreover, for a target node, since the influence of triangles of different distances is also distinct, we further introduce the definition of k-order triangles.  

\begin{definition} (k-Order Triangle): 
For node $v \in \mathcal{V}$, given a triangle $(v_0, v_1, v_2)$, if the shortest distance from these three nodes to $v$ is $k$, then we define this triplet as a k-order triangle of $v$. 
\end{definition}

Generally speaking, the triangle order $k$ is a natural number, which is smaller than the order of the neighborhood $K$. 
%For a target node, the k-order triangle is more relevant with it if k is smaller. 

\section{Observation}

\begin{figure}[!htbp]
  \centering
  \vspace{-10pt}
  \subfigure{
    \label{fig:obser1}
    \includegraphics[width=0.25\linewidth]{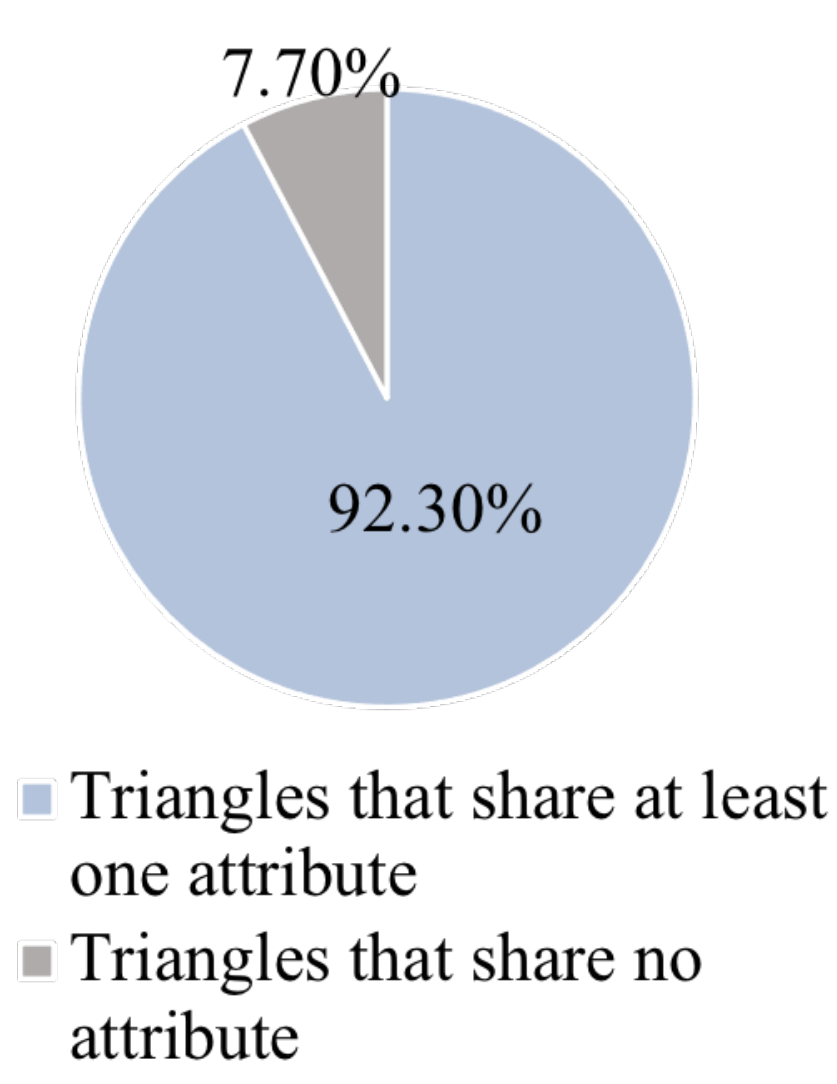}
 }
  \subfigure{
    \label{fig:obser2}
    \includegraphics[width=0.68\linewidth]{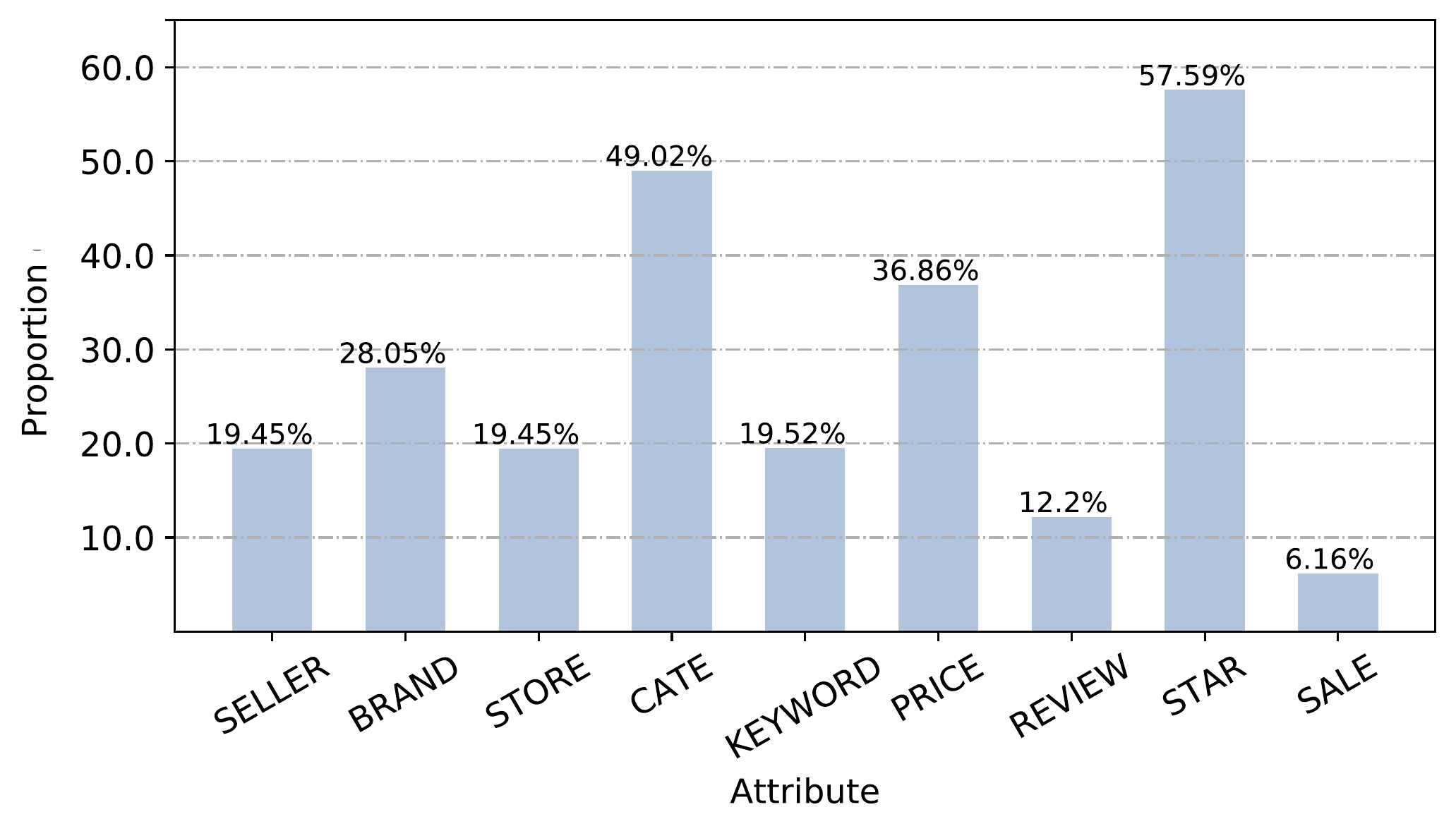}
 }
 \vspace{-10pt}
 \caption{Observation of triangles. The left figure shows most of triangles have three nodes sharing the same attributes. The right figure indicates the sharing attributes of different triangles are distinct.}
 \vspace{-5pt}
 \label{fig:obser}
\end{figure}

In this section, we observe the triangles in item-item co-occurrence graphs and summarize their properties by statistics and analysis. 
Specifically, we use the Alibaba dataset (more details about the dataset are in Table 1) to construct the graph and randomly select 100,000 items. 
For each item, a group of triangles in its neighborhood is extracted. 
We compare multiple attributes of three items involved in each triangle. 
The attributes include brand, seller, store, category, and keyword, price, review, star, and sale. 
We counted how many triangles share attributes and which attributes are shared. 
The results are showed in Figure \ref{fig:obser}. 
Consequently, we summarize two properties of triangles in the item co-occurrence graph.

\textbf{(1) Intra-triangle homophily}
As showed in the left of Figure \ref{fig:obser1}, we find that 92.30\% triangles have the three items that share at least one attribute. 
%For example, a triangle contains three books written by the same writer or sold in the same store. 
Conversely, only 7.7\% of the triangles have the items that are not related clearly. 
Hence, we can summarize that three items in a triangle usually share some common attribute. 
%We assume that those shared attributes, such as category and store, can uncover the user's real motivations to click these items. 
%For example, maybe a user prefers the books of William Shakespeare, not all tragic novels. 
%Therefore, we regard triangles as the most basic user interest unit. 
%Compared with modeling single items, modeling triangles is more helpful to capture users' elusive and implicit interests. 

\textbf{(2) Inter-triangle heterophily}
In the right of Figure \ref{fig:obser2}, we counted the proportion of the triangles sharing each attribute. 
More than 49\% of the triangle belong to the same category, and only 6\% of the items in the triangles are of the similar sales. 
Therefor, although these triangles come from the neighborhoods of the same batch of items, the shared attributes of different triangles differs greatly. 
%Therefore, they reflect different aspects of user interests.
%A variety of triangles can introduce novel and diverse commodities to users. 
%As the result, users can obtain more exploration opportunities and expand their interests. 
%For example, we can recommend not only books of the same author or store, but also other books of the similar price or review star to the users.

\begin{table}[!htbp]
%\small
\center
\caption{Analysis of K-cliques.}
\vspace{-10pt}
\label{tab:clique}
{\begin{tabular}{c|c|c}
\toprule
\  & Occurrence Probability & Homophily \\
\hline
2-Clique    & 3.92\%    & 0.817  \\
3-Clique    & 1.77\%    & 0.909  \\
4-Clique    & 0.8\%     & 0.228  \\
5-Clique    & 0.21\%    & 0.005  \\
\bottomrule
\end{tabular}}
\vspace{-10pt}
\end{table}

So far in this paper, we have only focused on triangles, which are a particular kind of motifs (3-clique). 
Hence, we also study 2/4/5-cliques in a similar way. 
We extract a subgraph, randomly sample 2/3/4/5 nodes from it, and check if they can form a 2/4/5-clique. 
This process is repeated 1,000,000 times to calculate the occurrence probability of each clique. 
The probability can reflect the number of these cliques in the graph. 
In addition, we compare all the attributes of the items in each clique and measure the average homophily of cliques.
The results are shown in Table \ref{tab:clique}. 
We infer that, there are too few 4-cliques and 5-cliques to provide sufficient information while the homophily of 2-cliques is relatively lower due to the contingency.
Therefore, the proposed method targets on triangle structures.
It is worth mentioning our method can be easily extended to other kinds of motifs.

\section{Method}

\subsection{Efficient Triangle Extraction}
Extracting triangles is a time-consuming process, especially for large-scale graph data. However, as a real-time task, CTR prediction has a high demand for efficiency. 
Empirically, for a specific item, the triangles near to it have a stronger influence. Hence, in our paper, triangles are extracted only from small-scale neighborhood of each item. 
Take the 0-order triangle as an example, for each node in the co-occurrence graph, we sample all its 1-order neighbors. 
Then we check if each pair of neighbor nodes can form a triangle with the specific item. 
That is, it is necessary to judge whether there is an edge between each pair. 
We solve it by constructing a Hashmap for edges, which can limit the time complexity to $\mathcal{O}$(1). 
But it requires high storage capacity and encounters an obstacle when handling billions of edges. 
Thus, we use Bloom Filters \cite{broder2004network} to alleviate this problem. 
We can produce all the 1-order triangles after deduplication. 
Based on this, triangles of any order can be obtained in a similar way. 

\subsection{Diverse Triangle Selection}
Inspired by \cite{chen2018fast}, we adopt Determinantal Point Processes (DPP) to model the selection probability of a relevant yet diverse set of triangles. 
Specifically, a DPP on the triangle set $\mathcal{T}$ is a probability distribution on the powerset of $\mathcal{T}$. 
If it assigns non-zero probability to the empty set $\emptyset$, there exists a positive semi-definite kernel matrix $\mathbf{L} \in \mathbb{R}^{N \times N}$, such that for each subset of $\mathcal{S} \subseteq \mathcal{T}$, the probability of $\mathcal{S}$ is defined as follows:
%\vspace{-4mm}
\begin{equation}
{\small
\begin{split}
p(\mathcal{S})=\frac{\operatorname{det}\left(\mathbf{L}_{\mathcal{S}}\right)}{\operatorname{det}(\mathbf{L}+\mathbf{I})},
\end{split}
}
\end{equation}
where $\mathbf{L}_{\mathcal{S}}$ is $\mathbf{L}$ restricted to those rows and columns which are indexed by $\mathcal{S}$, and $\mathbf{I}$ is the identity matrix. 

As revealed in \cite{chen2018fast}, the kernel matrix $\mathbf{L}$ can be written as a Gram matrix $\mathbf{L}=\mathbf{T T}^{\top}$.
%, where $\mathbf{B} \in \mathbb{R}^{N \times d}$ and $d \ll N$
The rows of $\mathbf{T}$ are vectors representing the feature of triangles. 
Each row vector $\mathbf{T}_{i}$ are empirically constructed as the product of the triangle relevance score $r_{i}$ and the triangle feature vector $\mathbf{x}_{i} \in \mathbb{R}^{1 \times d}$, i.e., $\mathbf{T}_{i}=$ $r_{i} \mathbf{x}_{i}$. 
Hence, an element of $\mathbf{L}$ is written as $\mathbf{L}_{i j}=\mathbf{T}_{i} \mathbf{T}_{j}^{\top}=r_{i} r_{j} \mathbf{x}_{i} \mathbf{x}_{j}^{\top}$. 
Moreover, if the feature vector $\mathbf{x}_{i}$ is normalized, the cosine similarity between two triangles can be calculated as $\mathbf{C}_{i j}=\mathbf{x}_{i} \mathbf{x}_{j}^{\top}$. 
Hence, $\mathbf{L}$ can be re-written as follows:
%\vspace{-4mm}
\begin{equation}
{\small
\begin{split}
\mathbf{L}=\operatorname{Diag}\{\mathbf{r}\} \cdot \mathbf{C} \cdot \operatorname{Diag}\{\mathbf{r}\},
\end{split}
}
\end{equation}
where $\mathbf{C}$ is the triangle similarity matrix, and $\operatorname{Diag}\{\mathbf{r}\}$ is a diagonal matrix with the $i^{t h}$ element being $r_{i}$. 
Following \cite{chen2018fast}, we modify the DPP kernel matrix as follows: 
%\vspace{-4mm}
\begin{equation}
{\small
\begin{split}
\mathbf{L}=\operatorname{Diag}\{\exp (\alpha \mathbf{r})\} \cdot \mathbf{C} \cdot \operatorname{Diag}\{\exp (\alpha \mathbf{r})\},
\end{split}
}
\end{equation}
where $\alpha=\theta /(2(1-\theta))$ and $\theta \in(0,1)$. 
$\theta$ is a hyper-parameter used to trade-off between relevance and diversity. 
%For simplicity, we model the relevance score of each triangle using a linear function of its features as follows:
%$$
%r_{i}=\mathbf{a x}_{i}^{\top}
%$$
%where $\mathbf{a} \in \mathbb{R}^{1 \times d}$ is the model parameter. 
%As the triangle similarity matrix $\mathbf{C}$ is fixed when the triangle feature vectors are given, $\mathbf{a} \in \mathbb{R}^{1 \times d}$ becomes the only model parameter needs to be learned to determine the DPP kernel. 

In practice, we calculate the triangle feature vector by averaging the features of three nodes within it. 
As to the triangle relevance score, we introduce the inner weight and the outer weight of the triangle. 
The inner weight is the average weight of its three edges.
For the outer weight, we set the denominator as the sum of the distances from three inner nodes to the central node, and set the numerator as the sum of the edge weights connecting three inner nodes and the central node. 
Based on this, the triangle relevance score is the arithmetic square root of the product of the inner and outer weights.
After the DPP kernel $\mathbf{L}$ is determined, the inference algorithm \cite{chen2018fast} is applied to find the optimal diverse while relevant triangle set for the recommendation.

\subsection{Model Architecture}
\subsubsection{Embedding Layer}%%%%%%%%%%%%%%

 \begin{figure*}[!htbp]
     \centering
     \includegraphics[width=0.8\textwidth]{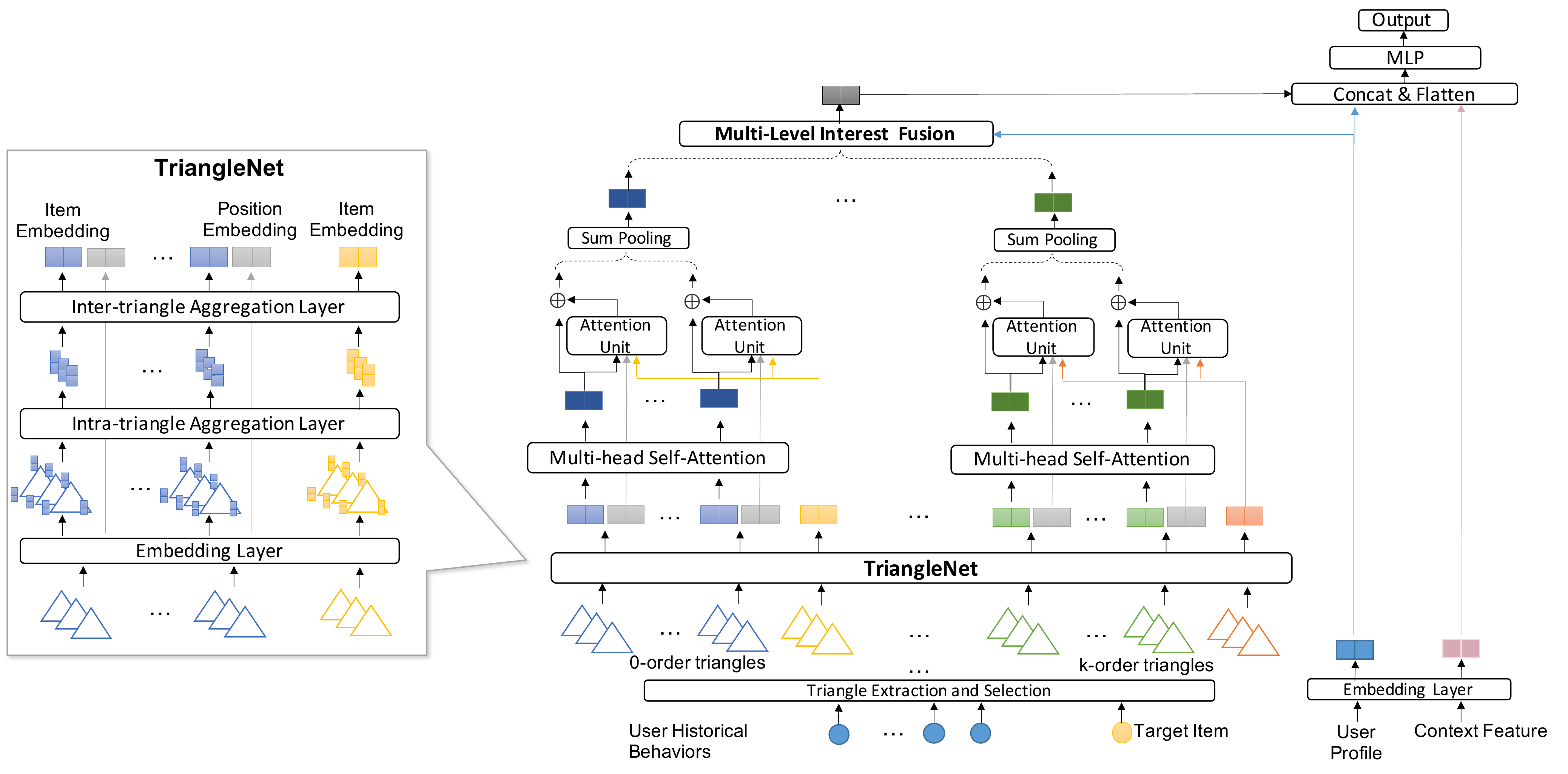}
     \vspace{-10pt}
     \caption{The model architecture of TGIN. Firstly, we introduce diverse and related triangles from the item-item graph to represent potential user interest units. Then, user behavior representations are enriched by intra-triangle and inter-triangle aggregation. Finally, triangles of different orders help to capture multi-levels of user interests, thereby improving the performance of click-through rate prediction.} 
     \label{fig:model}
     \vspace{-10pt}
 \end{figure*}

In the real-world recommendation scenario, both users and items have extensive features.
In this paper, we use four categories of feature: User Profile, User Historical Behavior, Context and Target Item. 
They are represented as $\mathbf{x}_{u}, \mathbf{x}_{b}, \mathbf{x}_{p}, \mathbf{x}_{c}$ and $\mathbf{x}_{t}$, respectively.
%User Profile contains features related to the user, e.g. user id, country and so on. Target Item refers to the candidate item with corresponding features such as item id, category id, statistical ctr and so on. User Historical Behavior is a list of user interacted items by clicking, purchasing or add-to-cart. Each item in this list has same feature fields as Target Item. Context is a group of features including but not limited to time, match type, trigger id and so on.  
Especially, there are several items in User Historical Behavior. Thus, it can be represented by $\mathbf{x}_{b} \in \mathbb{R}^{L \times d}$, where $L$ is the length of user behavior sequences and $d$ is the dimension of item embeddings. 
%Note that items in User Historical Behavior and Target Item share the same embedding matrices. 
%For example, the item id can be represented by a matrix $\mathbf{E} \in \mathbb{R}^{K \times d_{v}}$, where $K$ is the total number of items and $d_{v}$ is the embedding size with $d_{v} \ll K$. 
Besides, to model the dynamic changes of user interests, we also consider the position of each item in user behavior sequences. 
%Similar with the representations of item id $\mathbf{E}$, the position is represented by a matrix $\mathbf{P} \in \mathbb{R}^{K \times d_{p}}$ and $d_{p}$ is the dimension of the position embeddings.  
User Behavior Position is represented by $\mathbf{x}_{p}=\left\{\mathbf{p}_{1}, \mathbf{p}_{2}, \ldots, \mathbf{p}_{L}\right\} \in \mathbb{R}^{L \times d}$, where $\mathbf{p}_{L} \in \mathbb{R}^{d_{p}}$ is the position encoding for the $t$-th item. 
After triangle extraction and selection, we produce a set of triangles for all items in the user behavior sequence. 
We stack the node representations in all triangles.  
Consequently, User Behavior Triangle is represented as $\mathbf{Tr}_b \in \mathbb{R}^{L \times n \times 3 \times d}$ where $n$ is the number of triangles. 
Particularly, we also extract triangles for target items. 
Target Item Triangle is represented as $\mathbf{Tr}_t \in \mathbb{R}^{n \times 3 \times d}$. 
Notably, for the convenience of description, we ignore the order of the triangle here. 
In practice, we also consider triangles of different orders. 
We will discuss it later.
%$k$ is the maximum order of the triangles and 

\subsubsection{Intra-triangle Aggregation Layer}
Given a set of relevant and diverse triangles for user behaviors and target item, we use these triangles to represent the potential interest units behind items. 
Here, we use a simple pooling operation to aggregate intra-triangle information and produce the representations of user behavior sequences. 
%\vspace{-4mm}
\begin{equation}
{\small
\begin{split}
\mathbf{H}_b = \text{Average}(\mathbf{Tr}_b),
\end{split}
}
\end{equation}
where $\mathbf{H}_b \in \mathbb{R}^{L \times n \times d}$ and $\text{Average}$ indicates the average pooling operation. $n$ is the number of triangles for each item.
In particular, we also enrich the features of the candidate items by introducing triangles in a similar way. The representation of the target item is calculated as:
%\vspace{-4mm}
\begin{equation}
{\small
\begin{split}
\mathbf{H}_t = \text{Average}(\mathbf{Tr}_t),
\end{split}
}
\end{equation}
where $\mathbf{H}_t \in \mathbb{R}^{n \times d}$. Instead of modeling each behavior simply, incorporating triangles can provide opportunities to discover serendipitous items. 

\subsubsection{Inter-triangle Aggregation Layer}
In this part, we describe how to aggregate multiple triangles with multi-head self-attention \cite{vaswani2017attention, devlin2018bert} for each item a user clicks. 
We employ multi-head self-attention to realize the aggregation among multiple triangles. 
%Self-attention is a special attention mechanism, which has been successfully applied to a variety of tasks \cite{vaswani2017attention, devlin2018bert}. 
The input of the self-attention module consists of query, key, and value and these three components are the same. 
Multi-head self-attention is a combination of multiple self-attention structures, which can learn the relationship in different representation subspaces \cite{devlin2018bert}. 
To be specific, the output of the $\text{head}_{h}$ is calculated as follows:
%\vspace{-4mm}
\begin{equation}
{\small
\begin{split}
\text{head}_{h} &=\operatorname{Attention}\left(\mathbf{H}_b \mathbf{W}_{h}^{Q}, \mathbf{H}_b \mathbf{W}_{h}^{K}, \mathbf{H}_b \mathbf{W}_{h}^{V}\right) \\
&=\operatorname{Softmax}\left(\frac{\mathbf{H}_b \mathbf{W}_{h}^{Q} \cdot\left(\mathbf{H}_b \mathbf{W}_{h}^{K}\right)^{\top}}{\sqrt{d_{h}}} \cdot \mathbf{H}_b \mathbf{W}_{h}^{V}\right),
\end{split}
}
\end{equation}
where $\mathbf{W}_{h}^{Q}, \mathbf{W}_{h}^{K}, \mathbf{W}_{h}^{V} \in \mathbb{R}^{d_{h} \times d_{h}}$ are projection matrices of the
$h$-th head for query, key and value respectively. $d_{h}$ is the dimension of each head. Thus, each head represents a latent item representation in a subspace.

Then vectors of different heads are concatenated to produce the item representations, which can be defined as follows:
\begin{equation}
{\small
\begin{split}
\left.\mathbf{Z}_b = \text{MultiHead} \left(\mathbf{H}_{b}\right) = \text {Concat(head}_{1}, \text{head}_{2}, \ldots, \text {head}_h\right) \mathbf{W}^{O},
\end{split}
}
\end{equation}
where $h$ is the number of heads, $\mathbf{W}^{O}$ is a linear matrix. 

For the target item, we also need to aggregate multiple triangles to get its representation. Similarly, we use multi-head self-attention to get the target item representation:
\begin{equation}
{\small
\begin{split}
\mathbf{Z}_{t} = \text{MultiHead} \left(\mathbf{H}_{t}\right).
\end{split}
}
\end{equation}
Unlike previous works that used single item features, we aggregate multiple triangles to represent each item's potential interests. As a result, the model can determine whether the user behaviors closely match the target item.

\subsubsection{Behavior Refiner Layer}
In practice, the downstream task benefits from high-quality item representations. 
Hence, after obtaining the item representations, we describe how to further refine it. 
Inspired by \cite{xiao2020deep}, we use another multi-head self-attention same as the previous part to realize the interaction among multiple interests:
\begin{equation}
{\small
\begin{split}
\operatorname{head}_{h}^{\prime} &=\operatorname{Attention}\left(\mathbf{Z}_b \mathbf{W}_{h}^{\prime Q}, \mathbf{Z}_b \mathbf{W}_{h}^{\prime \mathbf{N}}, \mathbf{Z}_b \mathbf{W}_{h}^{\prime}\right) \\
&=\operatorname{Softmax}\left(\frac{\mathbf{Z}_b \mathbf{W}^{\prime}{h}_{h}^{Q} \cdot\left(\mathbf{Z}_b \mathbf{W}^{\prime}{ }_{h}^{\mathbf{K}}\right)^{\top}}{\sqrt{d_{h}}} \cdot \mathbf{Z}_b \mathbf{W}_{h}^{\prime}\right). 
\end{split}
}
\end{equation}
Similarly, $\mathbf{W}_{h}^{\prime}, \mathbf{W}_{h}^{\prime K}, \mathbf{W}_{h}^{\prime V} \in \mathbb{R}^{d_{h} \times d_{h}}$ are projection matrices of the $h$-th head for query, key and value respectively. We pack all output head vectors $\{{(head}_{1}^{\prime}, \text{head}_{2}^{\prime}, \ldots, \text {head}_h^{\prime} \}$ as a matrix $\mathbf{B} \in \mathbb{R}^{L \times h \times d}$.

An auxiliary loss \cite{xiao2020deep} is used to supervise better item representation learning. It uses the $(i+1)$ -th user behavior original item embedding, $\mathbf{z}_{i+1}$, to supervise the $i$ -th learnt item representation, $\mathbf{b}_{i}$. 
The positive example represents the actual next behavior, while the negative example is drawn from the entire item set, excluding the clicked items.
Mathematically, the auxiliary loss is formulated as:
\begin{equation}
{\small
\begin{split}
L_{aux}=-\frac{1}{N}\left(\sum_{i=1}^{N} \sum_{t} \log \sigma\left(\mathbf{b}_{t}^{i} \odot \mathbf{z}_{t+1}^{i}\right)+\log \left(1-\sigma\left(\mathbf{b}_{t}^{i} \odot \hat{\mathbf{z}}_{t+1}^{i}\right)\right)\right),\end{split}
}
\end{equation}
where $\sigma$ is the sigmoid activation function and $\odot$ denotes the inner product. $\hat{\mathrm{z}}_{t+1}^{i}$ is the original embedding of negative example, $N$ represents the number of training examples.

\subsubsection{User Behavior Modeling Layer}
In this paper, to capture the drifting trend of user interests over time, we extract position features from user behavior sequences. 
A position-aware attention mechanism is proposed to calculate the importance of user behaviors at different positions for user modeling. 
Specifically, the attention coefficient for every user behavior of user representations is calculated as follows:
\begin{equation}
{\small
\begin{split}
{\mathbf{c}} &=\operatorname{POS\_Attention}\left({\mathbf{Z}}_{t} {\mathbf{W}_{p}}^{Q}, {\mathbf{B}} {\mathbf{W}_{p}}^{K}, {\mathbf{x}}_{p} {\mathbf{W}_{p}}^{V}\right) \\
&=\operatorname{Softmax}\left(\mathbf{Z}_{t} {\mathbf{W}_{p}}^{Q} \odot \sigma \left({\mathbf{B}} {\mathbf{W}_{p}}^{K} + {\mathbf{x}}_{p} {\mathbf{W}_{p}}^{V}\right)\right),
\end{split}
}
\end{equation}
where ${\mathbf{W}_{p}}^{Q}, {\mathbf{W}_{p}}^{K}, {\mathbf{W}_{p}}^{V} \in \mathbb{R}^{d \times d}$ are projection matrices for query, key and value respectively. Thus, ${\mathbf{c}}$ represents a vector of the attention coefficients. Note that the position of each item in the user behavior is the reverse sequential order sorted by timestamp when it has occurred.
%, i.e. the more recent occurred behavior items will be put at the higher ranks. 

Considering the different importance of each behavior, we can obtain users representations for a target item. Mathematically, the adaptive representation of the user behavior w.r.t. the target item is calculated as follows:
\begin{equation}
{\small
\begin{split}
\mathbf{v} =\sum_{j=l}^{L} \mathbf{c}_{j} \mathbf{b}_{j},
\end{split}
}
\end{equation}
where $\mathbf{b}_{j} \in \mathbb{R}^{d}$ is the $j$-th row of $\mathbf{B}$. It represents the representation vector of the $l$-th item in the user behavior sequences. $\mathbf{c}_{j} \in \mathbb{R}$ denotes how relevant of the behavior $\mathbf{b}_{j}$ to the target item.

\subsubsection{Multi-Level Interest Fusion Layer}
Triangles of different orders reflect different levels of interest. 
Therefore, as mentioned earlier, for user historical behaviors and target items, we extracted the multi-order triangles. 
For each order $k$, the representation of user behaviors can be obtained in a similar way, which is denoted as $\mathbf{v}_k$. 
Considering their different importance to users, we fuse multi-level interests with the attention mechanism. 
We stack all the user behavior representations of different orders as $\mathbf{V}$.
The attention coefficient of the behavior representation $\mathbf{V}$ to the user can be calculated as:
\begin{equation}
{\small
\begin{split}
\mathbf{e} = \operatorname{Softmax}\left(\mathbf{V}\mathbf{W} \odot \mathbf{x}_{u}\mathbf{W}\right),  
\end{split}
}
\end{equation}
where $\mathbf{W}$ is a linear transformation matrix. Based on this, we can finally fuse multi-level interests for the learned representations. Mathematically, the adaptive representation of the user behavior sequences w.r.t. the user is calculated as follows: 
\begin{equation}
{\small
\begin{split}
\hat{\mathbf{x}}_{b}=\sum_{j=k}^{K} \mathbf{e}_{k} \mathbf{v}_{k}, 
\end{split}
}
\end{equation}
where $\mathbf{v}_{k} \in \mathbb{R}^{d}$ represents the representations learned from $k$-order triangles. 
$\mathbf{e}_{k} \in \mathbb{R}$ denotes how important the representation $\mathbf{v}_{k}$ is to the user.

%\hat{\mathbf{x}}_{b}

\subsubsection{Optimization}
All the features vectors, $\mathbf{x}_{u}, \hat{\mathbf{x}}_{b}, \mathbf{x}_{c}$ and $\mathbf{x}_{t}$, are concatenated and fed into the MLP layer for final prediction. Since the click-through rate prediction task is a binary classification task, the loss function is chosen as cross-entropy loss, which is usually defined as:
\begin{equation}
{\small
\begin{split}
L_{\text {target }}=-\frac{1}{N}\left(\sum_{i=1}^{N} y_{i} \log f\left(\mathbf{x}_{i}\right)+\left(1-y_{i}\right) \log \left(1-f\left(\mathbf{x}_{i}\right)\right)\right), 
\end{split}
}
\end{equation}
where $\mathbf{x}_{i}=\left(\mathbf{x}_{u}, \hat{\mathbf{x}}_{b}, \mathbf{x}_{c}, \hat{\mathbf{x}}_{t}\right) \in \mathcal{D}, \mathcal{D}$ is the training set with size $N$
$y_{i} \in\{0,1\}$ is the click label, $f(x)$ is the prediction output of our network. As we use the auxiliary loss to supervise item representation refining, the final loss can be defined as,
\begin{equation}
{\small
\begin{split}
L_{\text {total }}=L_{\text {target}}+\lambda L_{\text {aux }}, 
\end{split}
}
\end{equation}
where $\lambda$ is a hyper-parameter in order to balance the two sub-tasks.
\section{Experiment}
In this section, extensive experiments are conducted on two public datasets and an industrial dataset to verify the effectiveness of {\short}. 

\subsection{Dataset}

\begin{table}[!htbp]
%\small
\vspace{-10pt}
\center
\caption{Statistics of the datasets.}
\vspace{-5pt}
\label{tab:Datasets}
\begin{tabular}{c|c|c|c|c}
\toprule
Dataset & \#User & \#Item & \#Categoriy & \#Sample  \\
\hline
Books & 603,668 & 367,982 & 1,600 & 603,668\\
Electronics & 192,403 & 63,001 & 801 & 192,403 \\
Industrial & 9 million & 62 million & 7,131 & 1 billion \\
\bottomrule
\end{tabular}
\vspace{-5pt}
\end{table}

We adopt both public and industrial datasets. The statistics of all datasets are shown in Table \ref{tab:Datasets}:

\begin{itemize} 
\item {\bf Public dataset}: 
Following the setting of \cite{zhou2019deep}, we use two subsets of Amazon dataset \cite{mcauley2015image}: Books and Electronics. 
For these two datasets, we regard the reviews as the user behaviors and sort the reviews from each user by the time order. 
Based on user historical behaviors, our purpose is to predict whether a user will write a review. 
\item {\bf Industrial dataset}: We collect the click logs for 30 days from Alibaba online display advertising system. 
A sample instance consists of the candidate item, user, context and its historical behaviors. 
The items that user clicked in the previous 29 days contribute to user sequential behaviors. 
Our goal is to predict whether the user clicks at the candidate item on the last day.
\end{itemize}

\begin{table}[!htbp]
%\small
\vspace{-5pt}
\center
\caption{Results (AUC) on the public datasets.}
\label{tab:PublicResult}
\begin{tabular}{c|c|c}
\toprule
Model & Books & Electronics \\
\hline
%BaseModel           & 0.7686 $\pm$ 0.00253 & 0.7435 $\pm$ 0.00128 \\
Wide\&Deep           & 0.7735 $\pm$ 0.00051 & 0.7456 $\pm$ 0.00127 \\
PNN                 & 0.7799 $\pm$ 0.00181 & 0.7543 $\pm$ 0.00101  \\
DIN                 & 0.7880 $\pm$ 0.00216 & 0.7603 $\pm$ 0.00028  \\
GIN                 & 0.7967 $\pm$ 0.00039 & 0.7684 $\pm$ 0.00174 \\
MIMN                & 0.8262 $\pm$ 0.00109 & 0.7563 $\pm$ 0.00083 \\
DIEN w/o AuxLoss      & 0.7911 $\pm$ 0.00150 & 0.7640 $\pm$ 0.00073 \\
DIEN                & 0.8453 $\pm$ 0.00476 & 0.7792 $\pm$ 0.00243 \\
DMIN w/o AuxLoss      & 0.8319 $\pm$ 0.00150 & 0.7553 $\pm$ 0.00108 \\
DMIN                & 0.8550 $\pm$ 0.00137  & 0.7778 $\pm$ 0.00110 \\
\hline
{\short} w/o AuxLoss  & \textbf{0.8808 $\pm$ 0.00112} & \textbf{0.7908 $\pm$ 0.00016} \\
{\short}            & \textbf{0.8934 $\pm$ 0.00103} & \textbf{0.7944 $\pm$ 0.00202} \\
\bottomrule
\end{tabular}
\vspace{-10pt}
\end{table}

\subsection{Baselines}
We compare {\short} with the following state-of-the-art approaches:
\begin{itemize} 
\item {Wide\&Deep} \cite{cheng2016wide} consists of two parts: its deep model is the same as Base Model, and its wide model is a linear model.
\item {PNN} \cite{qu2016product} uses a product layer to capture interactive patterns between interfield categories.
\item {DIN} \cite{zhou2018deep} uses the embedding product attention mechanism to learn the adaptive representation of user behaviors w.r.t. the target item.
\item {GIN} \cite{li2019graph} is the first to mine and aggregate the user’s latent intention on the co-occurrence item graph with graph attention technique. 
\item {DIEN} \cite{zhou2019deep} designs an auxiliary network to capture user’s temporal interests and proposes AUGRU to model the interest evolution. 
\item {MIMN} \cite{pi2019practice} is a multi-channel memory network to capture user interests from long sequential behaviors. 
\item {DMIN} \cite{xiao2020deep} models latent multiple user interests from user behavior sequences to improve the CTR task. 
\end{itemize} 

\subsection{Experiment Setting}
As introduced in Section 2, the item-item co-occurrence graph is constructed with a sliding window. 
To prevent information leakage, graph construction is only based on the train set for each dataset. 
The window size is set as 3 to avoid too extensive connections in the graphs. 
In our experiment, the dimension of the item and user embedding is set as 18, and the size of position embedding is set as 2. 
The learning rate is 0.001 and the batch size is 128. 
The maximum order of triangles is 2. 
The lengths of user behavior sequences vary in different datasets. 
Therefore, we set the maximum length to 10, 20, 15 for the Books dataset, the Electronics dataset, and the industrial dataset respectively. 
The evaluation metrics is Area Under the Curve (AUC). 
To ensure the fairness of comparison experiments, we reuse the metrics already reported in original papers or choose optimal hyper-parameters carefully after reproducing the code for different baselines. 

%It means the the probability that a randomly chosen positive example is ranked higher than a randomly chosen negative example, reflects the ranking ability of the model. 
%It is defined as follows:
%$$
%\mathrm{AUC}=\frac{1}{m^{+} m^{-}} \sum_{x^{+} \in D^{+}} \sum_{x^{-} \in D^{-}}(I(f(x^{+})>f(x^{-})))
%$$
%where $D^{+}$ is the collection of all positive examples, $D^{-}$ is the collection of all negative examples, $f(\cdot)$ is the result of the model's prediction of the sample $x$ and $I(\cdot)$ is the indicator function. 

\subsection{Results on Public Datasets.}
The results on the public datasets are shown in Table \ref{tab:PublicResult}. 
Compared with the benchmark algorithm, our proposed model achieves the highest AUC scores on both two datasets. 
Among all baselines, compared with traditional methods, Wide \& Deep and PNN, 
DIN and other models can achieve higher AUC scores obviously by capturing user interests. 
GIN outperforms DIN, demonstrating that introducing the item-item co-occurrence graph can alleviate the behavior sparsity problem.
Due to longer user behavior sequences in the Books dataset, MIMN performs better on this dataset than on the Electronic dataset. 
DIEN uses a specially designed AUGRU, thus, it is better at modeling the interest evolving process. 
As to DMIN, its advantages benefit from capturing multiple interests. Markedly, the auxiliary loss can bring a large improvement to all of DIEN, DMIN and our method. 
It is worth mentioning that our method can outperform all baselines even without the auxiliary loss. 
We own it to resolving the problems of elusive motivation and diversity restriction effectively.

\begin{table}[!htbp]
%\small
\center
\caption{Results (AUC) on the industrial dataset.}
\vspace{-5pt}
\label{tab:IndustrialResult}
\setlength{\tabcolsep}{7mm}{\begin{tabular}{c|c}
\toprule
Model & AUC \\
\hline
%BaseModel           & 0.6350  \\
Wide\&Deep          & 0.6575  \\
PNN                 & 0.6583  \\
DIN                 & 0.6613  \\
GIN                 & 0.6626  \\
DIEN                & 0.6658  \\
DMIN                & 0.6691  \\
\hline
{\short}                & \textbf{0.6732}  \\
\bottomrule
\end{tabular}}
\vspace{-10pt}
\end{table}

\subsection{Results on Industrial Dataset}
We further conduct the experiments on the industrial dataset of display advertisement. 
The results are shown in Table \ref{tab:IndustrialResult}. 
%Wide \& Deep and PNN obtain better performance than BaseModel. 
Different from only one category of goods in the public Amazon dataset, our industrial dataset contains various types of items.
Thus, it is much more challenging and complex. 
DIN and DIEN use the attention mechanism to capture the relevance of items, significantly improving performance.
Furthermore, when compared to GIN, DIEN, and MIMN, our proposed model outperformed them all.
Even when compared to DMIN, which performs the best on this dataset, {\short} achieves a 0.004 absolute AUC gain.
It is notable that in commercial advertising systems with hundreds of millions of traffics, 0.004 absolute AUC gain is significant and worthy of model deployment empirically. 
{\short} shows great superiority to better understand and make use of the characteristics of user behavior data.

\subsection{Ablation Study}
\begin{figure}[!htbp]
  \centering
  \vspace{-5pt}
  \subfigure[Book Dataset]{
    \label{fig:ablation1}
    \includegraphics[width=0.43\linewidth]{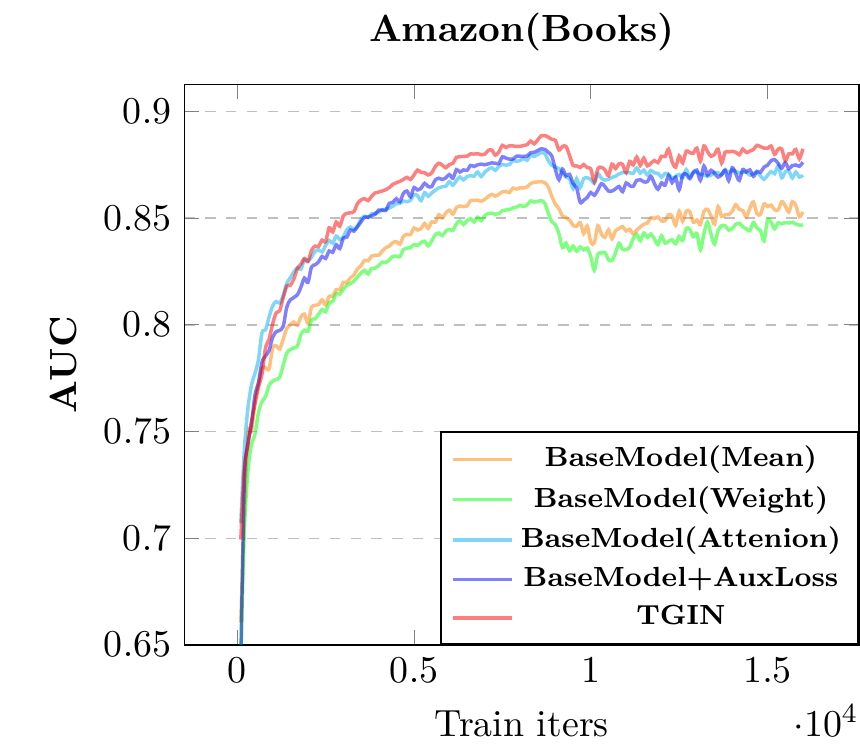}
 }
  \subfigure[Electronic Dataset]{
    \label{fig:ablation2}
    \includegraphics[width=0.45\linewidth]{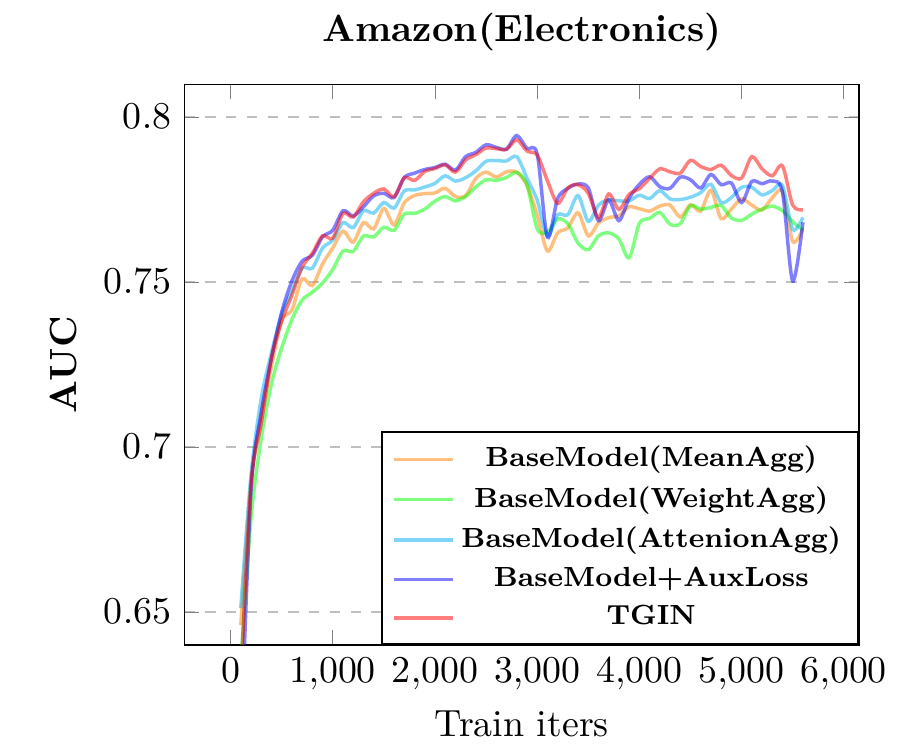}
 }
 \vspace{-10pt}
 \caption{The ablation study.}
 \vspace{-5pt}
 \label{fig:ablation}
\end{figure}

We test several variations of our full model, including BaseModel (MeanAgg), BaseModel (WeightAgg), BaseModel (AttentionAgg), BaseModel + AuxLoss.  
BaseModel is the base model to uses a set of triangles in the item-item co-occurrence graph to model every user behavior. 
There are three strategies to aggregate triangle information and click behaviors for each user, including MeanAgg, WeightAgg and AttentionAgg. 
They represent the average pooling, the weighted pooling and the attention pooling, respectively.
BaseModel + AuxLoss denotes the base model with the best pooling operation and the auxiliary loss.
All of these variations adopt the weight-based strategy to sample triangles. 
And {\short} is our full model, which samples relevant and diverse triangles  as introduced in Section 4.2.

Figure \ref{fig:ablation} shows the results of different variations of our full model on the public datasets. 
When compared to DIN in Table \ref{tab:PublicResult}, all triangle-based models achieve a clear advantage on the Books and Electronics datasets.
We own it to the effectiveness of modeling triangles as the basic units of user interests. 
It makes a great contribution to capturing implicit user interests. 
Considering different aggregation strategies, we find that the best choices for these two datasets are different: 
the attention pooling is better for Books while the weighted pooling is suitable for Electronics. 
We speculate that the reason is that the user behavior of the Electronic is sparser than that of Books, and complex aggregation mechanisms may cause overfitting. 
For all the experiments in this paper, we always choose the best aggregation strategies for each dataset.
As to the auxiliary loss, it brings great improvements for both datasets. 
It reflects the importance of supervision information for the learning of sequential interests. 
However, the improvement on Electronics is not as obvious as that on Books. The difference may derive from the longer user behavior sequences in the Books dataset. 
%Sequential information is more crucial for this dataset, which making it earn more from the auxiliary loss. 

\subsection{Hyperparameter Analysis}

\begin{figure}[!htbp]
  \centering
  \vspace{-10pt}
  \subfigure{
    \label{fig:parameter1}
    \includegraphics[width=0.45\linewidth]{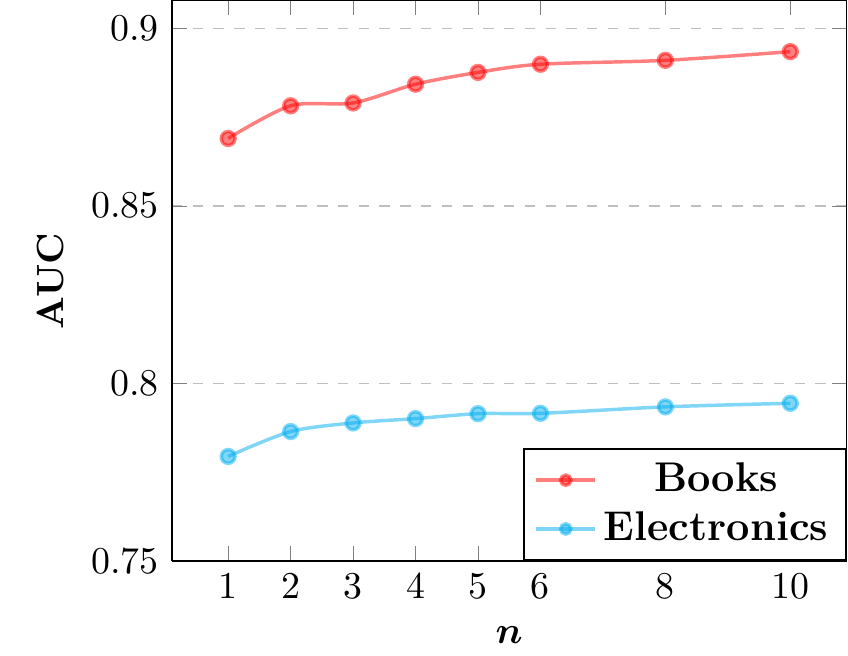}
 }
  \subfigure{
    \label{fig:parameter2}
    \includegraphics[width=0.45\linewidth]{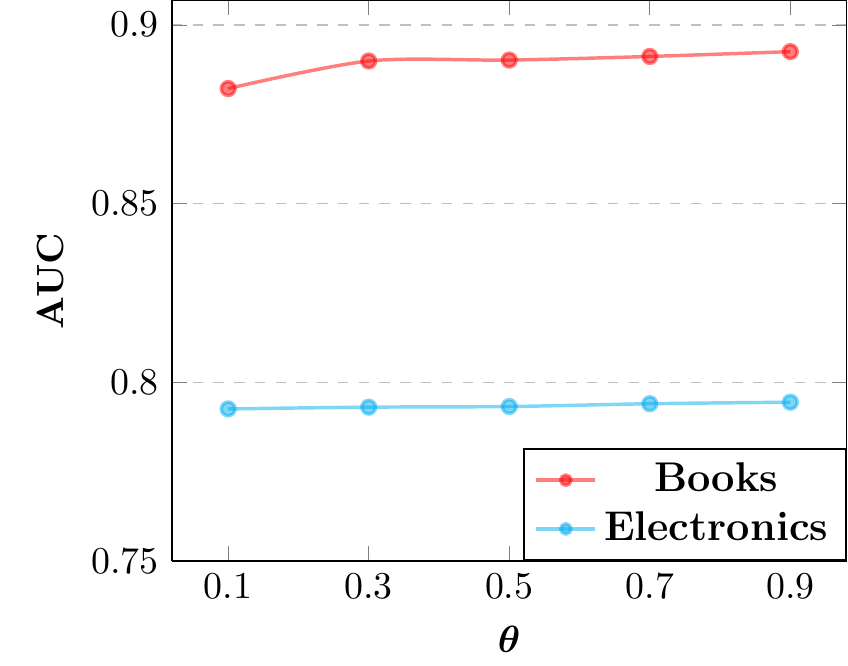}
 }
 \vspace{-5pt}
 \caption{Hyperparameter analysis.}
 \vspace{-5pt}
 \label{fig:parameter}
\end{figure}

Now we examine the influence of two key parameters in our
framework: the number of triangles $n$ and the parameter $\theta$ in triangle selection.
The results on two public datasets are shown in Figure \ref{fig:parameter}.
We can find that when the number of triangles increases from 1 to 10, our model can achieve better results on two datasets simultaneously.
Because more triangles can provide more clues about user interests.
However, in practice, we set $n$ as 10 to balance efficiency and effectiveness.
We set the parameter $\theta$ for a trade-off between relevance and diversity when choosing triangles. 
As shown in the right of Figure \ref{fig:parameter}, it has a little effect on the Electronics dataset while the Books dataset benefits from a higher $\theta$. 
We assume that the item features in these two datasets are very sparse, which may have a negative impact on triangle selection.

\subsection{Diversity Analysis}
\begin{figure}[!tbp]
     \centering
     \includegraphics[width=0.5\linewidth]{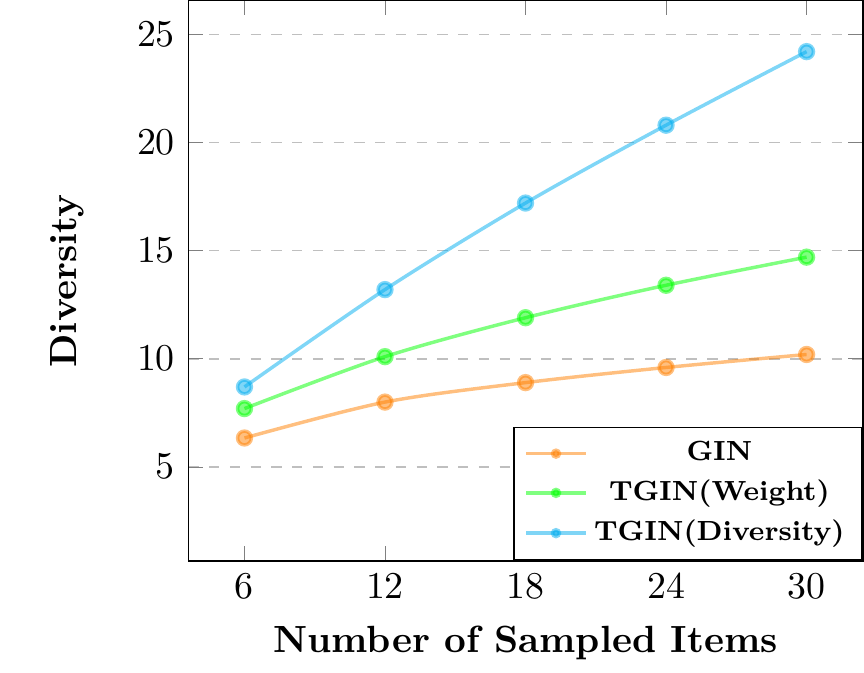}
     \vspace{-5pt}
     \caption{Diversity analysis.}
     \label{fig:diversity}
 \end{figure}

To verify whether a variety of triangles can bring more novel items, we analyzed different item sampling methods, including weight-based node sampling in GIN, weight-based triangle sampling in TGIN, and diversity-based triangle sampling in TGIN (in Section 4.2). 
For the fairness of comparison, we use different methods to sample the same number of items on the industrial dataset, and then, count the number of keywords involved in these items. 
The more different keywords, the more diverse items. 
The experimental results are shown in Figure \ref{fig:diversity}. 
We can find that the items sampled by GIN are relatively homologous. Because GIN is bias to popular or similar items.
In contrast, by introducing triangle structures, weight-based sampling in GIN can improve diversity to a certain extent. 
In our full model, we try to select diverse and relevant triangles.
In this way, higher diversity can be achieved with fewer items. 
Inspired by this, we improve the model efficiency by reducing the number of sampled triangles.

\subsection{Model Efficiency}
The calculation time of TGIN is compared with other popular sequential models, including DIN, GIN, MIMN, DIEN, DMIN. 
We recorded the training time of these models per epoch on the Books dataset with 1 NVIDIA Tesla P100 GPU. 
For TGIN, we extracted and selected triangles through preprocessing. 
The order of triangles is fixed to 2.
The number of triangles is set as 5. 
For other methods, we directly run their released codes or reproduce their results. 
The experimental results show that the efficiency of TGIN is comparable to, if not better than, that of these baselines. 
Specifically, DIN costs 367 seconds, TGIN 430s, GIN 457s, DMIN 1,309s, DIEN 1,354s, and MIMN 7,954s per epoch. 
GIN and TGIN are slower than DIN because they introduce item-item co-occurrence graphs as an auxiliary for broader exploration.
When compared with other baselines, TGIN shows a distinct advantage by modeling a few and diverse triangles. 
For online deployment, we optimized every component for our algorithm and finally obtain quick responses to user clicks for online advertising. More details are described in the next section.

\subsection{Result from online A/B testing.}

 \begin{figure}[!tbp]
     \centering
     \includegraphics[width=0.8\linewidth]{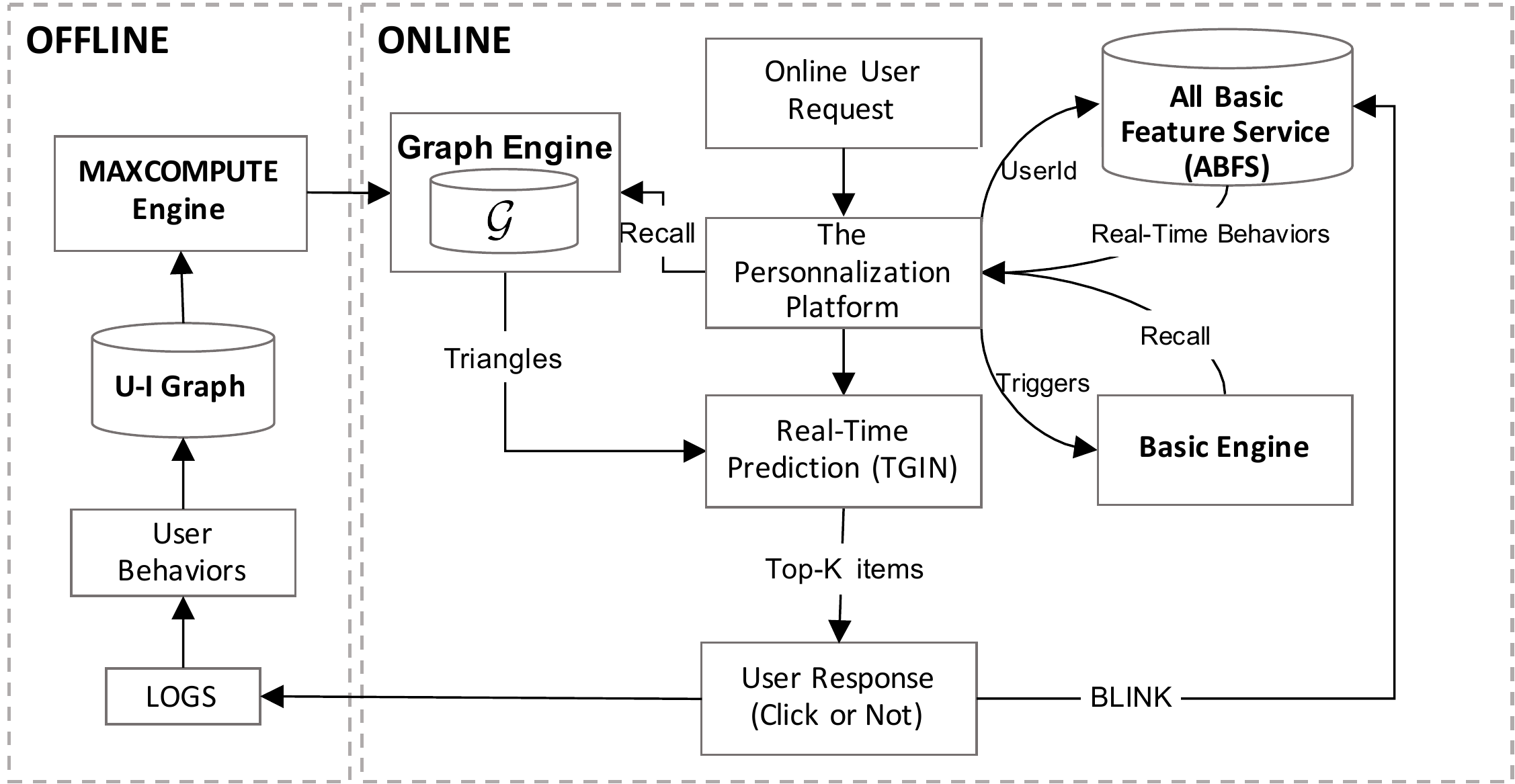}
     %\vspace{-5pt}
     \caption{TGIN deployment for online CTR.}
     \label{fig:system}
     \vspace{-5pt}
 \end{figure}
 
From April 10 to April 17 in 2021, careful online A/B testing was conducted in the merchant advertising system of AliExpress (one of the biggest global e-commerce platform).
During almost a week’s testing, TGIN has improved UV L-P by 3.62\% and L-GMV by 3.16\% compared to the base model. 
Besides, TGIN has improved commission earn per user (EPU) by 3.20\%. 
This is a significant improvement and demonstrates the effectiveness of TGIN. 
Now, as showed in Figure \ref{fig:system}, TGIN has been deployed online and serves the merchant traffic, which contributes a significant business revenue growth.

In practice, we employ the TPP platform to handle real-time user requests. 
ABFS (Ali Basic Feature Service) is used to obtain real-time behaviors and user portraits. 
Then we obtain the recall candidates via BE (Basic Engine). 
In the following step, item attribute information and the corresponding triangle index is requested via iGraph (GRAPH ENGINE). 
Finally, we calculate the CTR score through the TGIN model service deployed by RTP (REAL-TIME PREDICTION) and recommend items to users for online display. 
The average response time is about 30 milliseconds, which is sufficient to meet the timeliness requirement of the CTR prediction task.
Meantime, the user behavior log is returned to ABFS through FLINK real-time data.
In particular, all the triangle indexes are established offline. 
Firstly, a large-scale item-item co-occurrence graph is built for AliExpress. 
%It contains 60 million nodes and 3 billion edges. 
%The traditional triangle extraction algorithm is difficult to apply on such a large graph. 
The triangle extraction and selection algorithm is implemented using Alibaba MaxCompute and MapReduce. 
%We produce triangle indexes of about 1 billion triangles. 
To improve the efficiency, we restrict the number of triangles strictly.
%The DPP algorithm is used to sample a small but diverse set of triangles. 
%Finally, for each item, we sample 5 triangles.

\section{Related Work}

In recent years, click-through rate prediction for online advertising
has become one of the hot research topics. 
Inspired by the success of deep learning in computer vision \cite{chen2020simple} and natural language processing \cite{devlin2018bert}, deep learning-based methods have been proposed to improve the performance of the click-through rate prediction task \cite{song2016multi, covington2016deep, yu2016dynamic, zhou2018deep, zhou2019deep, zhou2018atrank, feng2019deep, li2019graph, xiao2020deep, DBLP:conf/wsdm/ZhangQCLLZMC21, DBLP:journals/apin/LiDZWW20, DBLP:conf/wsdm/0002PZKFL21, DBLP:conf/sigir/HuangHTCQCL21, DBLP:conf/sigir/FeiZZZQL21, DBLP:conf/sigir/ChengX21}. 
As the pioneer works, YoutubeNet \cite{covington2016deep} and DeepFM \cite{guo2017deepfm} present their potential of modeling complex interaction between users and items. 
Later, DIN \cite{zhou2018deep} learns adaptive representations of historical behavior sequences to infer user preference.
Inspired by DIN, the majority of subsequent works adopt this type of paradigm, such as DIEN \cite{zhou2019deep}, SDM \cite{lv2019sdm}, DSIN \cite{feng2019deep}, MIMN \cite{pi2019practice}, and DMIN \cite{xiao2020deep}.
%DIEN \cite{zhou2019deep} and SDM \cite{lv2019sdm} devote to capturing users temporal interests and modeling their sequential relations. 
%DSIN \cite{feng2019deep} focuses on capturing the relationships of users’ inter-session and intra-session behaviors. 
%MIMN \cite{pi2019practice} and HPMN \cite{ren2019lifelong} apply the neural turing machine to model users’ lifelong sequential behaviors. 
%DMIN \cite{xiao2020deep} models latent multiple user interests from user behavior sequences to improve the CTR task. 
Though these methods bring great improvements, they summarize user interests only based on historical behaviors. 
It suffers greatly from the sparsity of user behaviors. 
Besides, they cannot jump out of user specific historical behaviors so that users lose their opportunities for possible interest exploration. 

Some researchers attempt to integrate graph structures for enriching relationships among user behaviors and items \cite{wang2018billion, li2019graph, feng2020mtbrn}. 
GIN \cite{li2019graph} is the first work of end-to-end joint training of graph learning and CTR prediction tasks. 
It introduces item-item co-occurrence graphs to enrich user historical behaviors. 
Typically, they sample neighbors with higher weights of clicked items in the graph. 
Then they model user behaviors by aggregating information from these neighbors. 
Also, some works introduce knowledge graphs to explores their potential interests, such as RippleNet \cite{wang2018ripplenet}, KGAT \cite{wang2019kgat}, ATBRG \cite{feng2020atbrg} and MTBRN \cite{feng2020mtbrn}. 
However, most of these existing works still lack an effective mechanism to capture implicit user interests. 
In addition, due to the weakness of their sampling strategy from graphs, they are more biased towards the popular or similar commodities.  
Therefore, it is difficult to make diverse recommendations beyond user existing behaviors.

\section{CONCLUSION}

In this paper, we propose a novel and effective framework named {\ours} ({\short}) for the click-through rate prediction task.
For each clicked item in user behavior sequences, we introduce triangles in its neighborhood of the item-item graph as a supplement.
{\short} regards these triangles as basic units of user interests, which provide the clues to capture the real motivation for a user clicking an item. 
We characterize user behaviors by aggregating information of several interest units to alleviate the elusive motivation problem. 
By selecting diverse and relative triangles, {\short} brings in novel and serendipitous items for users to break diversity restrictions and expand exploration opportunities. 
\section{Acknowledgement}
	This work is funded in part by the Natural Science Foundation of China Projects No.U1936213, No.U1636207.

\bibliographystyle{ACM-Reference-Format}
\balance
\bibliography{0_main}

\end{document}